\documentclass[preprint,review,12pt]{elsarticle}

\usepackage{amsmath,amssymb,amsfonts}
\usepackage{algorithmic}
\usepackage{booktabs}
\usepackage{makecell}
\usepackage[numbers,sort&compress]{natbib}
\biboptions{sort&compress}
\usepackage{amsfonts}
\usepackage{graphicx}
\usepackage{textcomp}
\usepackage{gensymb}
\usepackage{hyperref}
\hypersetup{pdfauthor=whatever}
\usepackage[section]{placeins}

\journal{Computerized Medical Imaging and Graphics}

\begin{document}

\begin{frontmatter}



\title{QUIZ: An Arbitrary Volumetric Point Matching Method for Medical Image Registration}

\author[label1,label2]{Lin Liu \corref{cor2}}
\ead{lin.liu2@siat.ac.cn}
\author[label1,label2]{Xinxin Fan \corref{cor2}}
\ead{xx.fan@siat.ac.cn}
\author[label3]{Haoyang Liu}
\ead{16620777686@163.com}
\author[label1]{Chulong Zhang}
\ead{cl.zhang1@siat.ac.cn}
\author[label3]{Weibin Kong}
\ead{337174822@qq.com}
\author[label1]{Jingjing Dai}
\ead{jj.dai@siat.ac.cn}
\author[label3]{Yuming Jiang}
\ead{ymjiang2@stanford.edu}
\author[label1]{Yaoqin Xie}
\ead{yq.xie@siat.ac.cn}
\author[label1]{Xiaokun Liang \corref{cor1}}
\ead{xk.liang@siat.ac.cn}

\address[label1]{Shenzhen Institute of Advanced Technology, Chinese Academy of Sciences, Shenzhen, 518055, China}
\address[label2]{University of Chinese Academy of Sciences, Beijing, 100049, China}
\address[label3]{Guangdong Medical University, Dongguan, 523808, China}    
\address[label4]{Department of Radiation Oncology, Stanford University, Stanford, 94305, USA}

\cortext[cor1]{Corresponding author}
\cortext[cor2]{First Author and Second Author contribute equally to this work.}

\begin{abstract}
Rigid pre-registration involving local-global matching or other large deformation scenarios is crucial. Current popular methods rely on unsupervised learning based on grayscale similarity, but under circumstances where different poses lead to varying tissue structures, or where image quality is poor, these methods tend to exhibit instability and inaccuracies.

In this study, we propose a novel method for medical image registration based on arbitrary voxel point of interest matching, called query point quizzer (QUIZ). QUIZ focuses on the correspondence between local-global matching points, specifically employing CNN for feature extraction and utilizing the Transformer architecture for global point matching queries, followed by applying average displacement for local image rigid transformation.

We have validated this approach on a large deformation dataset of cervical cancer patients, with results indicating substantially smaller deviations compared to state-of-the-art methods. Remarkably, even for cross-modality subjects, it achieves results surpassing the current state-of-the-art.

Finally, our \href{https://github.com/louelin/QUIZ.git}{code and a portion of the data} will be available online.

\end{abstract}




\end{frontmatter}

\section{Introduction}
\label{}
Medical image registration is a crucial component of medical image analysis, with applications spanning a broad range of clinical scenarios, including image-guided preoperative planning, the construction of population-level atlases, and the diagnosis of various conditions \cite{haskins2020deep}, \cite{lesterSurveyHierarchicalNonlinear1999}. 

Traditional techniques for medical image registration rely on iterative optimization approaches to minimize an energy function \cite{sengupta2022survey}, \cite{shenImageRegistrationLocal2007}, \cite{zheng2021progressive} guiding the deformation field towards an optimal solution. 
Traditional techniques for medical image registration rely on iterative optimization approaches to minimize an energy function, guiding the deformation field towards an optimal solution. Notable traditional methods include Demons \cite{vercauteren2009diffeomorphic}, \cite{LORENZI2013470}, HAMMER \cite{1175091}, \cite{SHEN2009954},  Elastix \cite{klein2009elastix}, SyN \cite{avants2008symmetric}, and LDDMM \cite{beg2005computing}. 

Deep learning image registration (DLIR) has made significant strides in addressing minor or low-complexity deformations, particularly in the context of brain atlas development \cite{fanBIRNet}, \cite{MSnet}, \cite{RecursiveC}. The advent of large labeled datasets featuring local deformations within a few millimeters has led to marked improvements in brain registration accuracy \cite{abbas2022}. Nevertheless, when confronted with tasks involving more pronounced distortion, such as inter-patient abdominal magnetic resonance imaging (MRI) or inspiratory-to-expiratory computed tomography (CT) lung registration, DLIR performance lags behind that of traditional methods \cite{7892934}, \cite{hansen2020tackling}. For example, lung structures typically exhibit displacement deviations on the order of several centimeters \cite{Heinrich2019ClosingTG}, posing a formidable challenge to DLIR, which is better suited for less complex intra-patient registration. 

Radiation therapy, a prevalent treatment for cervical cancer, often comprises a combination of brachytherapy (BT) and external beam radiotherapy (EBRT) \cite{Ma2021DeepLA}. EBRT employs X-rays or alternative particle beams to target tumor tissue from outside the body, whereas BT involves positioning a radiation source in close proximity to the tumor tissue \cite{vordermark2016radiotherapy}. Accurate alignment is essential for precise reconstruction of the radiation source, accurate determination of tumor location, and preservation of healthy tissues due to the distinct irradiation positions and directions between EBRT and BT \cite{swamidas2020image}. Nonetheless, the DLIR method renders such alignment virtually unattainable because of the considerable discrepancies in scanning field of view and patient posture between EBRT and BT images. As the patient's condition evolves during BT treatment, the treatment plan must be updated accordingly \cite{Rigaud2019DeformableIR}. Tissue edema and other morphological changes can result in the radioactive source deviating from its anticipated position, thereby impacting the overall effectiveness of the therapy. However, due to the intricate and significant deformations caused by applicator insertion during BT, along with the gradual reduction of tumors throughout multiple treatments, registering EBRT and daily CT scans for BT at different time points presents a formidable challenge that demands innovative approaches \cite{Bondar2012IndividualizedNA}.

Specifically, the initial displacements of the scanned images for these two treatment modalities for cervical cancer vary widely and have high spatial complexity. However, existing methods perform poorly on the initial rigid transformation, which directly leads to failure of the subsequent elastic deformation, or even to unreasonable deformations such as overstretching in order to optimize in the wrong direction for a smaller similarity metric \cite{Bondar2010ASN}. Unfortunately, most of the current methods focus on the fine-structure correspondence of the elastic deformation and neglect the automatic alignment initialisation for large deformations \cite{r11,r12,r13}.

Our research has been focused on devising novel approaches to make more efficient use of relevant semantic information and incorporate additional constraints to improve the accuracy and robustness of medical image registration methods in cases involving significant deformations. In a departure from traditional DLIR methods, we present a novel transformer-based arbitrary correspondence point-matching network for medical image registration. The primary contributions of this work are outlined as follows:
\begin{itemize}
\item [1)]
We introduce a novel volumetric point-matching network grounded in the Transformer architecture, enabling arbitrary self-querying point correspondence and offering a wealth of potential applications in the medical image registration domain.
\item [2)]
In this study, we present a supervised registration technique that streamlines the matching of densely sampled points or feature points, obviating the need for an actual deformation field. Experimental findings indicate that our proposed method outperforms existing techniques when applied to images with large displacements, addressing a key limitation of current DLIR approaches.
\item [3)]
We have manually annotated corresponding anatomical landmarks on publicly available datasets and released them for scholarly access. This initiative is poised to significantly contribute to the advancement of similar future research endeavors in this domain.
\end{itemize}

\section{Method}
\subsection{Problem formulation}
Traditional image registration approaches involve iterative optimization of a similarity measure. Let $M, F \in R^{H \times W \times D}$ represent the moving and reference image volumes, respectively. The objective of image registration is to identify a deformation field $\phi$ that warps $M$ towards $F$, minimizing the dissimilarity between the warped image $M(\phi)$ and the fixed image $F$.

In this research, we introduce an innovative method that recasts volumetric medical image registration as a task concerning voxel coordinates $x \in D^{M}$, where $D^{M}$ encompasses all voxel coordinates in the reference image. Since our focus is on examining matching relationships between points on volumetric images, $M$ is set to $3$. $x^{t}$ denotes the coordinate of the anatomical structure in the search image that shares the precise physical location as $x$, corresponding to $x$'s location in the reference image. We define $\bar{x}=x^{t}-x, \bar{x} \in R^{M}$ to represent the displacement vector between the corresponding physical coordinate in the search image $I'$ and the point of interest in the reference image $I$. For $q \in R^{N \times M}$, where $N$ signifies the number of points of interest, and $q$ represents all query point coordinates, $N$ refers to several key feature points or a set of dense coordinate points randomly sampled on the surface, contingent upon specific application requirements.

Given a dataset $D$ containing pairs of images and corresponding point coordinates among them with identical anatomical information, our objective is:

\begin{equation}
\underset{\Phi}{\arg \min } \quad E_{\left(q, q^t, I, I^{\prime}\right) \sim D}=L_{\text {pair }}+\alpha L_{\text {trans }}
\end{equation}
with
\begin{equation}
L_{\text {pair }}=\left\|T_{\Phi}\left(q \mid I, I^{\prime}\right)+q-q^t\right\|_2^2
\end{equation}
In the search image $I'$, let $q^t$ denote the set of coordinates corresponding to $q$, and let $T$ represent our quizzer block, parameterized by $\Phi$.

\begin{equation}
L_{t r a n s}=-N C C
\end{equation}

The NCC is defined as follows:

\begin{equation}
\begin{aligned}
& \operatorname{NCC}\left(I, I^a\right)=\frac{\operatorname{Cov}\left(I, I^a\right)}{\sqrt{\operatorname{Var}(I) \operatorname{Var}\left(I^a\right)}} \\
& =\sum_{x \in \Omega} \frac{\sum_{x_i \in x}\left(I\left(x_i\right)-\bar{I}(x)\right)\left(I^a\left(x_i\right)-\bar{I}^a(x)\right)}{\sqrt{\sum_{x_i \in x}\left(I\left(x_i\right)-\bar{I}(x)\right)^2 \sum_{x_i \in x}\left(I^a\left(x_i\right)-\bar{I}^a(x)\right)^2}}
\end{aligned}
\end{equation}

where $x^a = Mx$, and let $M$ represent:

\begin{equation}
M=\left[\begin{array}{cc}
A & T_{\Phi}\left(q \mid I, I^{\prime}\right) \\
0 & 1
\end{array}\right]
\end{equation}

A is the identity matrix.

\begin{figure*}[ht]
\centerline{\includegraphics[width=1\columnwidth]{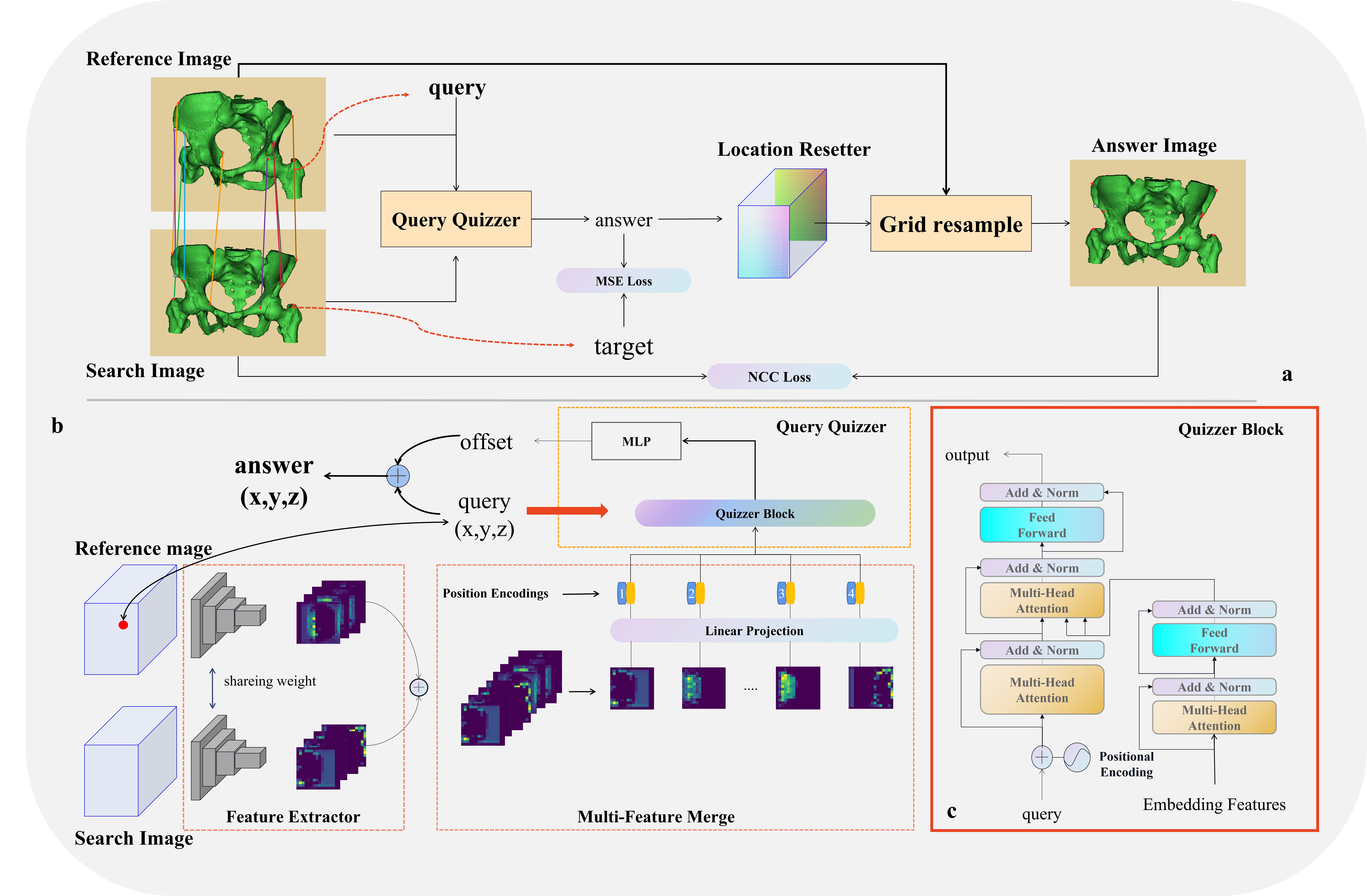}}
\caption{Structure of the proposed QUIZ for volumetric medical image registration: (a) Workflow for training strategies. (b) Overview of the QUIZ module. (c) Illustration of the quizzer block.}
\label{fig:fig1}
\end{figure*}

\subsection{Network architecture}

Figure \ref{fig:fig1}(a) illustrates the proposed QUIZ framework, which consists of three main components designed to perform matching across images: (1) an encoder that extracts features from the reference and search images using shared weights, (2) a quizzer block that interacts with both the image pairs and the points of interest, and (3) a position resetter that leverages deviations from the previous module's output to make positional corrections to the images.

Figure \ref{fig:fig1}(b) shows the feature extractor in our framework, which is based on the backbone of the ResNet10. Let $I_{r}, I_{s} \in \mathbb{R}^{D \times H \times W}$ represent the reference image and the search image, respectively, both of which are cropped and resized, with $D, H, W = 128$.

The feature representation for each image is extracted independently using the shared-weight encoder. These individual feature maps are then merged along the z-axis to produce the combined feature map $I_{c} \in \mathbb{R}^{C \times D_{c} \times H_{c} \times W_{c}}$, where $D_{c} = \frac{D}{8}$, $H_{c} = \frac{H}{8}$, $W_{c} = \frac{W}{4}$, and $C = 128$.

In the next step, the quizzer block in Figure \ref{fig:fig1}(c) is responsible for determining the displacement deviation of each interest point in the combined feature map $I_{c}$. This is achieved by interacting with both the image pairs and the points of interest. Once the deviations are computed, the average displacement in the $x, y$, and $z$ directions is calculated and used to represent the search image's overall direction.

Finally, we employ the grid sampling method on the search image to render the image differentiable, allowing for gradient-based optimization during training. The position resetter module then leverages the deviations computed by the quizzer block to make positional corrections to the images, resulting in an improved matching across the reference and search images.

\subsection{Baseline Methods}
In this study, we performed a comparative evaluation of our proposed QUIZ method alongside three other image registration techniques: the conventional symmetric image normalization method (SyN) \cite{avants2008symmetric}, and two unsupervised deep learning deformable registration methods, VoxelMorph \cite{r14} and TransMorph \cite{t04}. SyN is a well-regarded classical deformable registration algorithm, recognized for its efficacy across various applications, and is frequently employed as a benchmark for comparing novel methods. VoxelMorph, conversely, is a widely acknowledged unsupervised deep learning deformable registration method, often regarded as the pioneering and foundational unsupervised nonlinear registration algorithm. Lastly, TransMorph is a recently introduced unsupervised deformation registration algorithm, proven to be the contemporary state-of-the-art in medical image registration.

Throughout our experiments, we utilized the extensively adopted software package, Advanced Normalizing Tools (ANTs) \cite{ants}, for registering medical images, implementing SyN with a cross-correlation similarity measure and a maximum iteration set to $[150, 100, 50]$ for iterative optimization. For the learning-based approaches (VoxelMorph and TransMorph), we employed the official online implementations supplied by their respective authors to achieve optimal performance, training from the ground up and adopting NCC as the similarity loss and Dice loss as the auxiliary loss.

The chosen smooth regularization weight and learning rate were set at $0.01$ and $1e-4$, respectively. For equitable comparison, we utilized identical training, validation, and test sets, as well as seed points. Given the difficulty of aligning images with substantial deviations using learning-based approaches alone, we adopted two distinct methods: translation preprocessing and direct deep learning-based registration. ANTS translation was employed for pre-registration of image pairs.

\section{Experiments and Results}
\subsection{Datasets}
Our study utilized a dataset of 100 patients, each with one CT scan for EBRT and multiple CT images for BT. Specifically, the dataset contains 100 CT images for EBRT and 442 CT images for BT. For our experiments, we partitioned the patients into sets: $80\%$ for training, $10\%$ for validation, and $10\%$ for testing, ensuring a comprehensive evaluation of the performance and generalization of our proposed method. The search images were resampled to $512\times 512 \times 109$ with a spacing of $1.25mm \times 1.25mm \times 5mm$ according to the reference image space. We performed no additional image pre-processing and compared all benchmark methods under identical conditions to ensure a fair evaluation, thereby minimizing any biases or inconsistencies that might arise during the analysis.

We additionally performed validation in a publicly available dataset \cite{AYorke2021QualityAO}. The dataset was a retrospective study (2014-326) approved by the Beaumont Research Institutional Review Board of 58 patients selected for pelvic regional therapy at the Beaumont Radiation Oncology Centre. Each patient underwent a planning CT scan on a 16-row Philips Brilliance large-aperture CT scanner (Philips NA Corp, Andover, MA) covering the entire anatomical region with a fixation device. Each patient had a CBCT image acquired on an onboard imager of the Elekta Synergy® linear accelerator (Elekta Oncology Systems Ltd., Crawley, UK) for daily image guidance.The pixel range of the CBCT images ranged from $512 \times 512 \times 512$ to $512 \times 512 \times 110$, with pixel sizes from ( The CBCT images ranged from $512 \times 512 \times 88$ to $512 \times 512 \times 110$ pixels, 
with pixel sizes ranging from $(1 \times 1) \text{mm}^2 $ and 3-mm slice thickness to $(0.64 \times 0.64) \text{mm}^2 $ and 2.5-mm slice thickness. The planning CT was resampled to the same in-plane dimensions as the CBCT, and the image content was shifted to place the anatomical isocentre at the centre of the image volume.

We performed careful selection and manual labelling of anatomical marker points for CT-CBCT image pairs. On average, there were 154 landmark pairs (ranging from 91 to 212). These landmark pairs underwent a rigorous manual selection and expert-guided reorientation process. The selected landmark pairs are publicly available at \href{https://github.com/louelin/QUIZ.git}{https://github.com/louelin/QUIZ.git} with comprehensive instructions for use.

\begin{figure*}[!h]
    \centering
    \includegraphics[width=1\columnwidth]{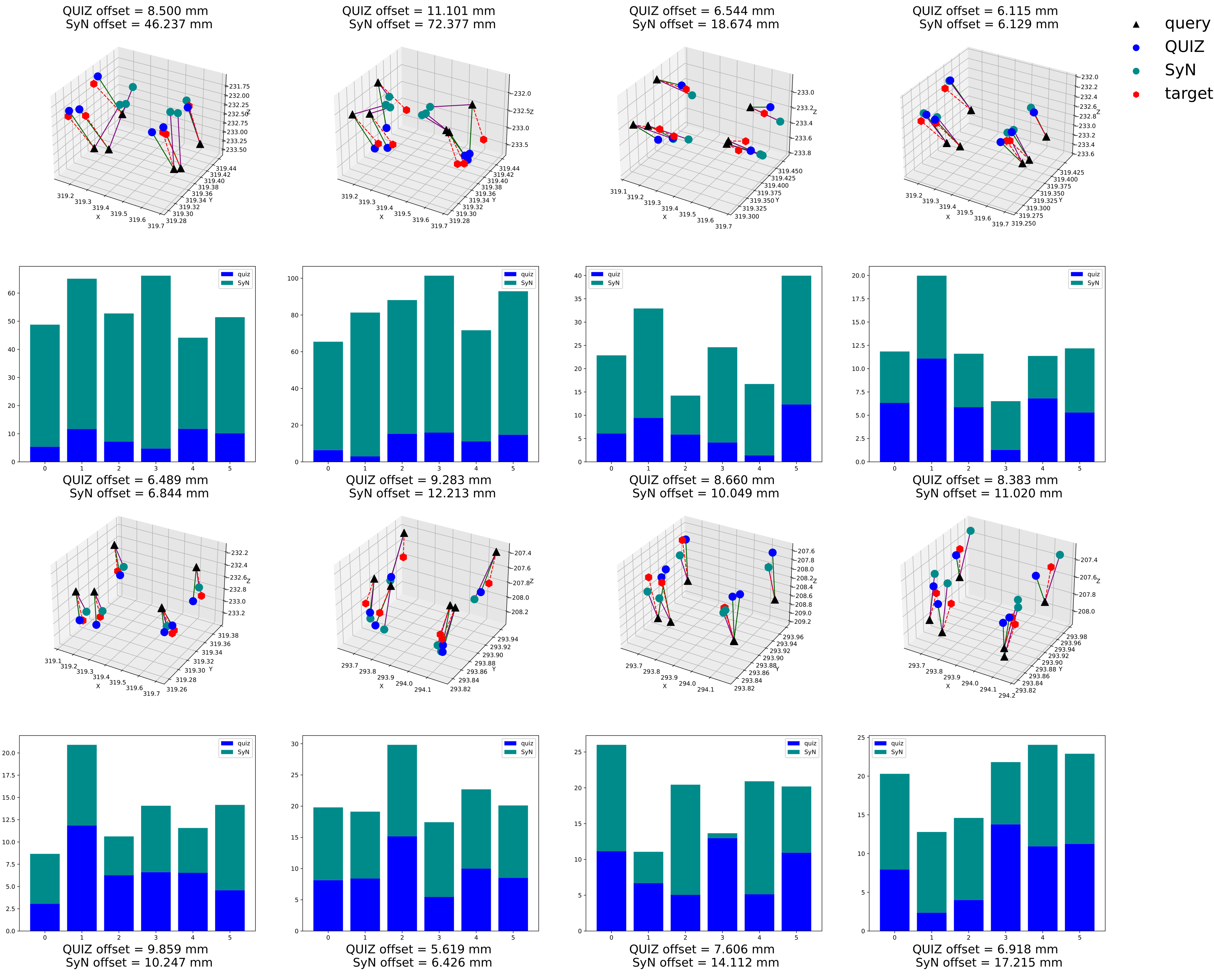}
    \caption{Visualizing the offsets for 8 test samples, each comprising six feature points, two distinct methods, QUIZ (blue) and SyN (cyan), are illustrated. The black triangle represents the query feature point, while the red dot signifies the anticipated corresponding point within the search image. The axes within the example denote the world coordinate values of the target points.}
    \label{fig:plteps}
\end{figure*}

\begin{table*}
	\setlength{\abovecaptionskip}{0cm}  
	\setlength{\belowcaptionskip}{0.2cm} 
	\centering
	\caption{Results for image registration across various cervical cancer exposure durations. Six expert-identified landmark pairs were employed to compute both the mean TRE in millimeters and the rTRE for each instance.}
	\resizebox{\textwidth}{!}{%
		\setlength{\tabcolsep}{7mm}
		\small
		\begin{tabular}{llllllllll}
			\toprule
			Experiment & TRE $(\mathrm{mm}) \downarrow$ & rTRE $\downarrow$\\
			\midrule
			SyN\cite{avants2008symmetric} & $60.268(57.665)$ & $0.0822(0.0787)$ \\
			VoxelMorph\cite{r14} &  $85.087(45.619)$ & $0.1214(0.0602)$ \\
			TransMorph\cite{t04} & $88.924(44.146)$ & $0.1214(0.0602)$ \\
			VoxelMorph+Translation & $62.280(59.230)$ & $0.0850(0.0808)$ \\
			TransMorph+Translation & $61.178(58.506)$ & $0.0833(0.0798)$ \\
			QUIZ & $\mathbf{33.235(14.623)}$ & $\mathbf{0.0453(0.0199)}$ \\
			\hline
		\end{tabular}%
	}
	\label{table1}
\end{table*}

\subsection{Evaluation Metrics and Results }
To evaluate the accuracy of our method, TREs were computed for six anatomical and pathological points in the pelvis using 230 pairs of daily CT scans designated for BT, annotated by radiologists. TRE was determined as the mean Euclidean distance between corresponding points before and after registration, providing a quantitative measure of registration performance:

\begin{equation}
\operatorname{TRE}(A, B)=\frac{1}{N} \sum_{i=1}^N\left\|a_i-b_i\right\|
\end{equation}

Here, $a_i$ and $b_i$ represent the $i$-th landmark coordinate vector in search image A and the corresponding answer point vector, respectively. Table \ref{table1} displays our experimental findings. Post pre-registration, both TransMorph and VoxelMorph exhibit performance comparable to traditional techniques, while our proposed method boasts nearly twice the effectiveness of alternative approaches. We have observed that, even after registration, the actual prediction accuracy for point pairs has yet to meet the standards for clinical use. However, it is imperative to emphasize that this is solely a study on rigid pre-registration aimed at addressing the challenge of local-global matching. Subsequent steps will necessitate deformable registration, which will further align the anatomical structures. We assessed the performance of our approach using the relative target registration error (rTRE), which accounts for the TRE normalized by the distance between anatomical landmarks, as calculated through the formula below:

\begin{equation}
\operatorname{rTRE}=\frac{T R E}{\sqrt{w^2+h^2+d^2}}
\end{equation}

where $w, h$, and $d$ represent the image dimensions in the $x, y$, and $z$ directions, respectively. This metric allows for a more comprehensive evaluation of the registration performance by accounting for the relative spatial relationships between landmarks.

Table \ref{table4} reveals the average displacement error derived from comparing the ground truth values of the anatomically marked points in the search image with the coordinates predicted by our proposed network. The time required to test a pair of scans is also tabulated. Fig. \ref{fig:plteps} visually portrays the matching relationship and error of the 3D feature points, substantiating the exceptional stability and accuracy attained by our QUIZ method. These results demonstrate the efficacy of our proposed method in comparison to conventional-based techniques, highlighting its potential for practical application in radiation therapy planning. In terms of computational efficiency, our method has demonstrated remarkable results. Using the same rigid registration approach, our method achieved nearly a 20-fold improvement in translation. Furthermore, it surpasses the performance of learning-based SyN\cite{avants2008symmetric} methods. It is noteworthy that the SyN\cite{avants2008symmetric} algorithm currently lacks GPU support, and the results in the table are based on CPU computations.

\begin{table}
	\setlength{\abovecaptionskip}{0cm}  
	\setlength{\belowcaptionskip}{0.2cm} 
	\centering
	\caption{Quantitative evaluation of the registration performance of BT and EBRT CT images encompasses registration accuracy (lower values indicate superiority) and speed (decreased values are preferable).}
	\setlength{\tabcolsep}{2mm}
	\normalsize
	\begin{tabular}{llllllllll}
		\hline Method & offset$(\mathrm{mm})$  & time $(\mathrm{s/item})$  \\
		\hline SyN\cite{avants2008symmetric} & $23.242 \pm 23.222$ & $157.706$  \\
		VoxelMorph\cite{r14} & $34.101 \pm 18.560$ & $1.038$  \\
		TransMorph\cite{t04} & $32.319 \pm 17.989$ & $0.914$  \\
		VoxelMorph+Translation & $24.041 \pm 23.847$ & $29.041$  \\
		TransMorph+Translation & $23.586 \pm 23.541$ & $28.917$  \\
		QUIZ& $\mathbf{11.947 \pm 5.837}$ & $\mathbf{1.203}$  \\
		\hline
	\end{tabular}%
	\label{table4}
\end{table}

In addition, we validated our method on top of publicly available datasets and the results are displayed in Table  \ref{table3}, illustrating that our method significantly outperforms traditional algorithms. Notably, the traditional state-of-the-art (SOTA) registration algorithm, SyN, is not as effective as the method calculating optimal translations. This is due to the fact that learning registration methods based on Normalized Cross Correlation (NCC) similarity measures are influenced by the incomplete information contained in local images. These learning methods, during the optimization process, forcibly stretch local images to maintain consistency in size with global images, without considering the structural correlation between the local and the overall image.

\begin{table*}[hb]
	\setlength{\abovecaptionskip}{0cm}  
	\setlength{\belowcaptionskip}{0.2cm} 
	\centering
	\caption{Quantitative assessment of the registration performance of abdominal CT and CBCT images.}
	\resizebox{\textwidth}{!}{%
		\setlength{\tabcolsep}{8mm}
		\small
		\begin{tabular}{llllllllll}
			\toprule
			Experiment & offset(mm) $\downarrow$ & TRE(mm) $\downarrow$ & rTRE $\downarrow$\\
			\midrule
			Translaion & $5.422$ & $11.650$ & $0.016$  \\
			SyN &  $5.560$ & $11.914$ & $0.016$ \\
			QUIZ & $\mathbf{3.951}$ & $\mathbf{8.100}$ & $\mathbf{0.110}$ \\
			\hline
		\end{tabular}%
	}
	\label{table3}
\end{table*}

\begin{figure*}[htb]
	\centerline{\includegraphics[width=1\columnwidth,height=0.45\textwidth]{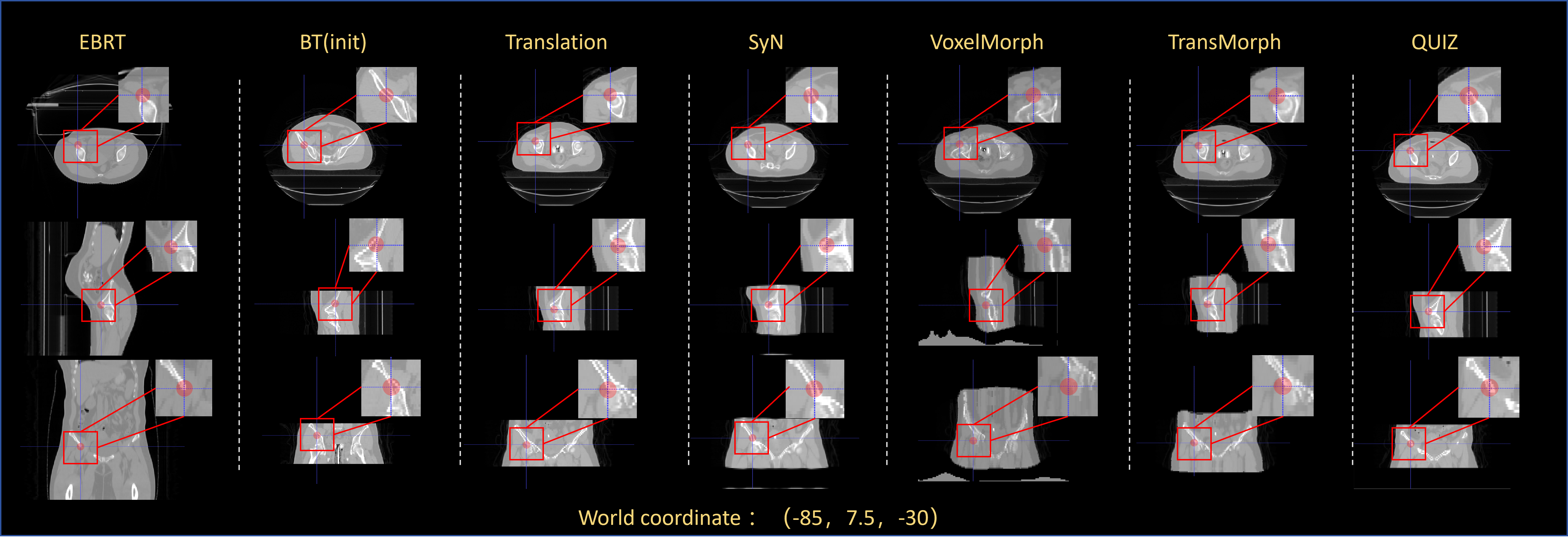}}
	\caption{Illustration of registration results for image pair example. Each row displays a specific orientation, while each column exhibits the registration result for a different approach. Our method is located in the last column, whereas the registration results depicted in the fifth and sixth columns have undergone a translation transformation. In order to show the difference in initial positions, we have chosen a fixed world coordinate. The overall registration results and zoomed-in details for each method are also plotted.}
	\label{fig:f3}
\end{figure*}

\subsection{Implementation Details}

For network parameter optimization during training, we utilized the Adam algorithm with a learning rate of $1e-4$, a batch size of $2$, and assigned $\alpha$ in loss to $0.01$. The model underwent training for approximately $13$ hours, with $2,000$ iterations executed. In real-time testing, the model's completion time was $1.22$ seconds, showcasing its efficiency and potential for rapid clinical decision-making.

All models were implemented on an Ubuntu system featuring an Intel $i5-12490F@4.0GHz \times 11$ processor and an NVIDIA GeForce RTX $A6000$ graphics card. The operational environment consisted of Python version $3.9.12$ and the PyTorch $1.12.1$ framework, with Cuda version $11.7$.

Given the limited collection of point pairs, we have augmented the data, incorporating various operations such as translation, scaling, random flipping, and coordinate axis swapping, while concurrently updating the landmark pairs.

Throughout the training procedure, we faced difficulties in attaining convergence during the initial phase when employing $L_{pair}$ and $L_{trans}$ co-training. Consequently, we initially trained the quizzer block and subsequently refined the network utilizing location resetter to achieve superior matching outcomes, thus overcoming the aforementioned challenges and enhancing the overall performance of our method.

\begin{figure*}[!htb]
	\centering
	\includegraphics[width=0.95\textwidth,height=0.55\textwidth]{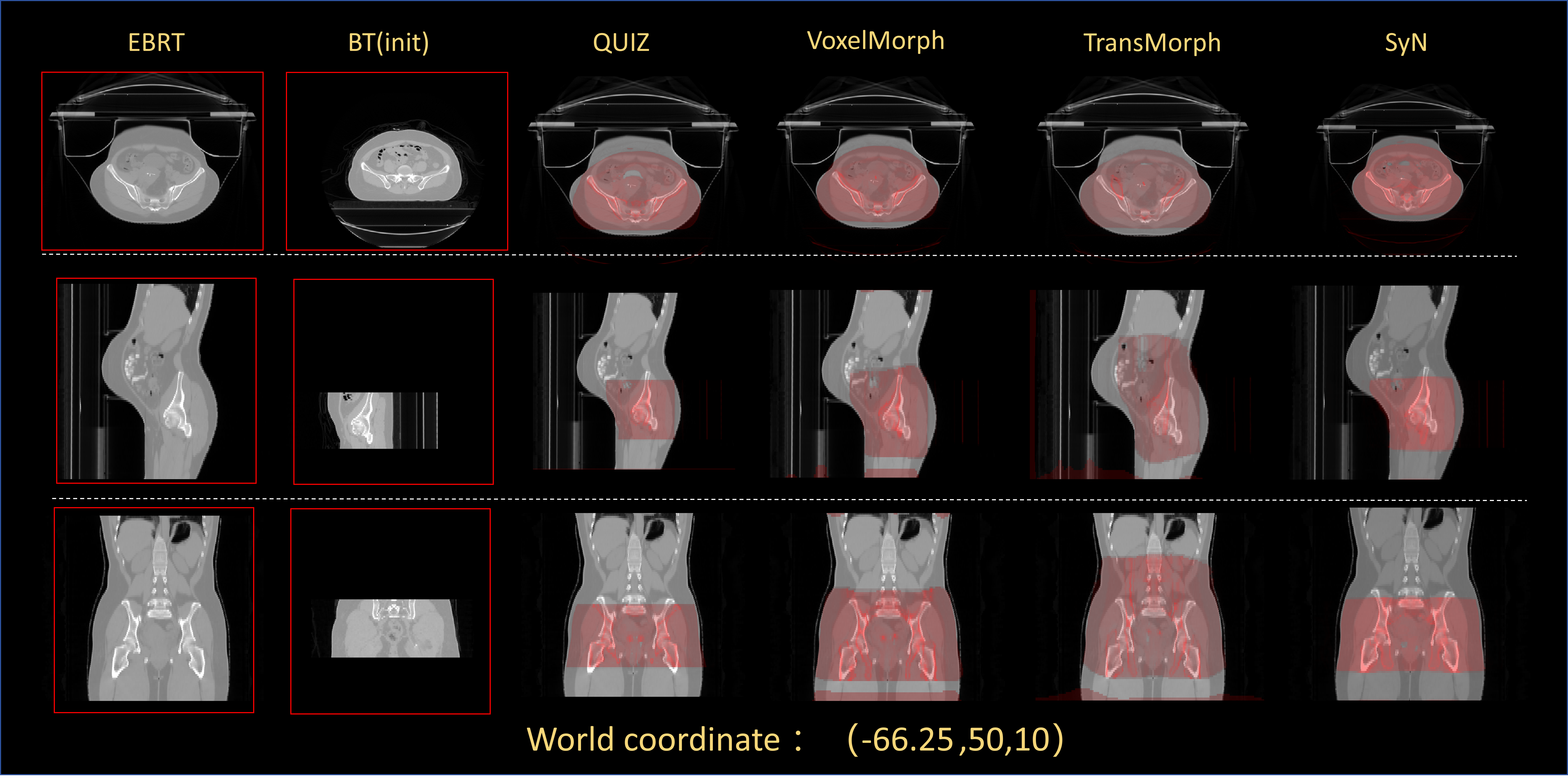}
	\caption{Example of registration results for EBRT and BT CT image pairs. The grey area represents the EBRT CT image, while the red area signifies the warped BT CT image. Our method is located in the third column.}
	\label{fig:f4}
\end{figure*}

\section{Disscussion}

\subsection{Registration Method Based on Grayscale Similarity Learning}

Pelvic region registration presents considerable challenges due to the pronounced variability in bladder volume, tumor size and shape, and complex deformation associated with BT applicator \cite{ghose2015review}. 
Based on the enlarged observation of the local details in the Fig. \ref{fig:f3}, we can draw the following conclusions: When employing learning algorithms to match search images with reference images, finding the optimal matching location becomes challenging, particularly when positional differences exist between them. Comparatively, the query point matching method we propose exhibits significant advantages in matching regions of interest. This method outperforms algorithms based on image grayscale similarity. For instance, substantial morphological differences exist in the abdominal regions when considering EBRT and BT, two distinct treatment modalities. These differences might lead to mismatches by algorithms reliant on image grayscale similarity, as they may incorrectly match images in different poses as pairs with smaller similarities. As illustrated in Fig. \ref{fig:f4}, algorithms involving geometric transformations like translation often yield noticeable biases. However, the query point matching method we propose is less affected by pose differences, accurately identifying matching points and effectively mitigating the impact between images with significant deformations.

\begin{figure*}[htb]
	\centering
	\includegraphics[width=1.0\textwidth,height=0.55\textwidth]{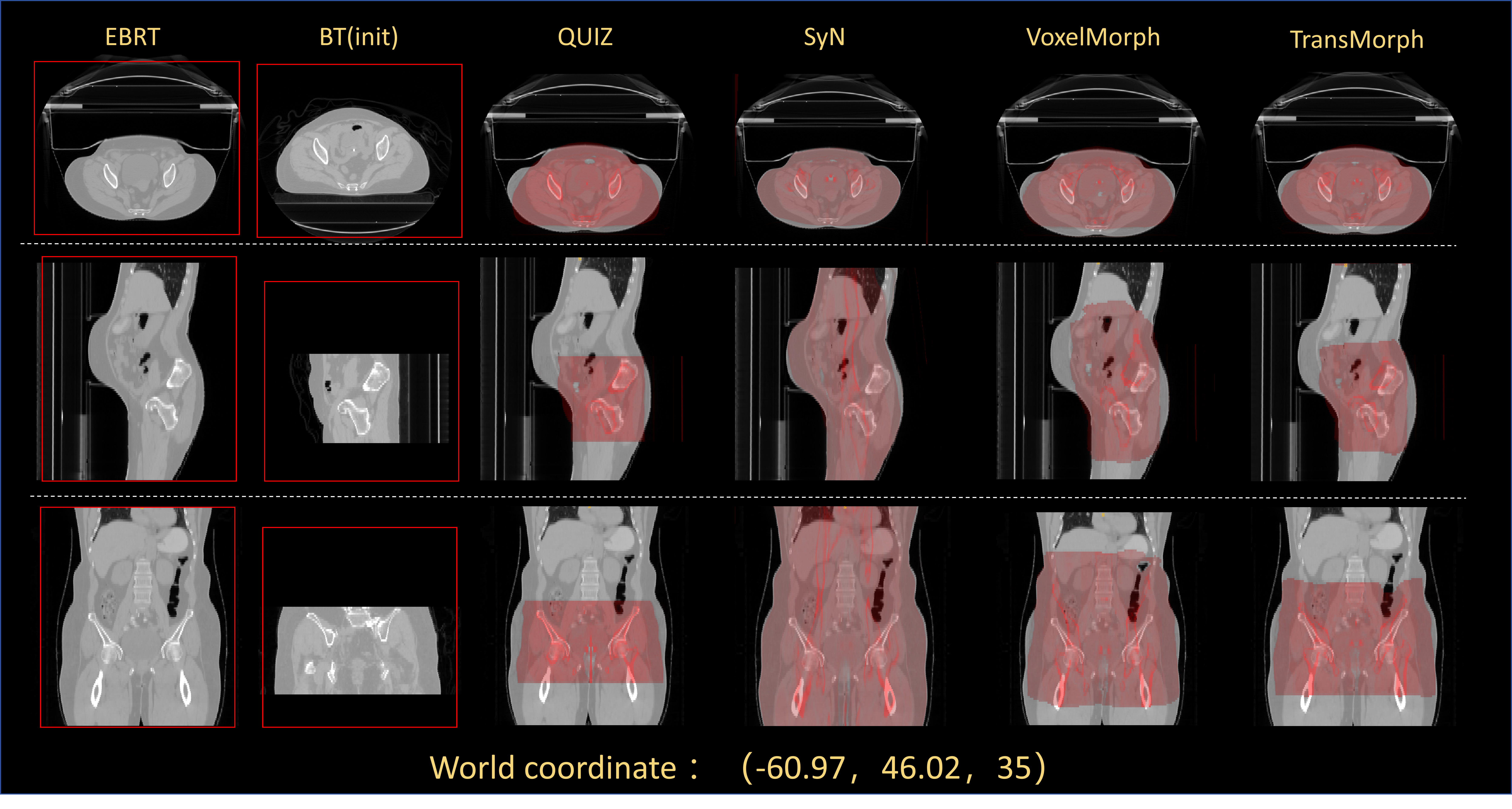}
	\caption{Another illustration showcasing the registration outcomes between EBRT and BT CT image pairs. The grey area represents the EBRT CT image, while the red area signifies the warped BT CT image. Our method is located in the third column.}
	\label{fig:f5}
\end{figure*}

\begin{figure*}[htb]
	\centering
	\includegraphics[width=1.0\textwidth]{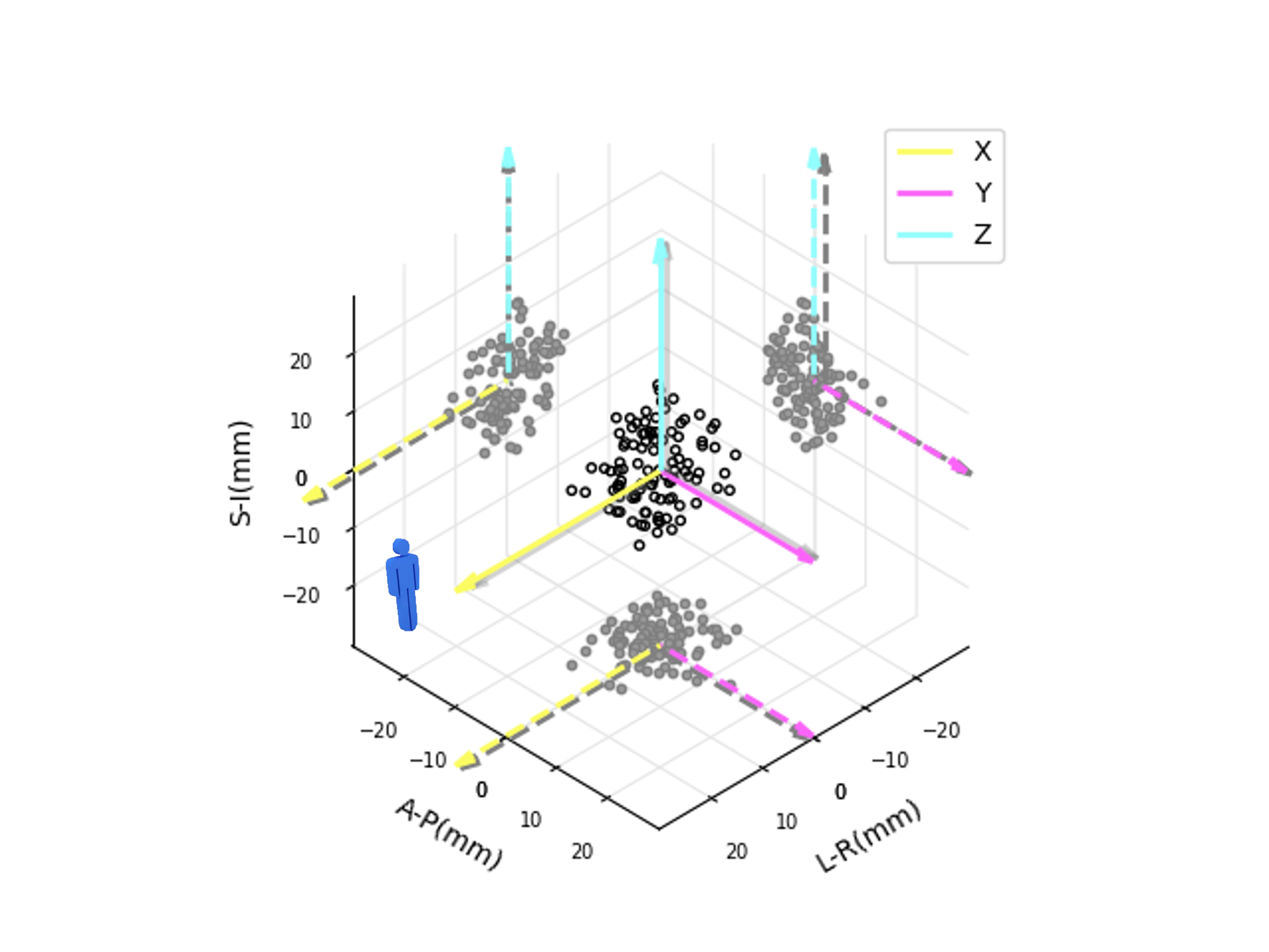}
	\caption{The disparities between the model predictions and target values within the publicly available cross-modality datasets, following data augmentation, were analyzed in distinct directions along the A-P (Anterior-Posterior), S-I (Superior-Inferior), and L-R (Left-Right) axes}
	\label{fig:f6}
\end{figure*}

During our experimental process, we observed results as depicted in Fig. \ref{fig:f5}. SyN, VoxelMorph, and TransMorph all induced stretching in the images, leading to structural distortion. This is attributed to the inability of learning algorithms based on image grayscale matching to adequately comprehend the relationship between local and global images. Intense stretching of local images to achieve higher similarity indices results in an imbalance in anatomical structure matching.

\subsection{Prediction Error in QUIZ}

\begin{figure*}[htb]
	\centering
	\includegraphics[width=1.0\textwidth]{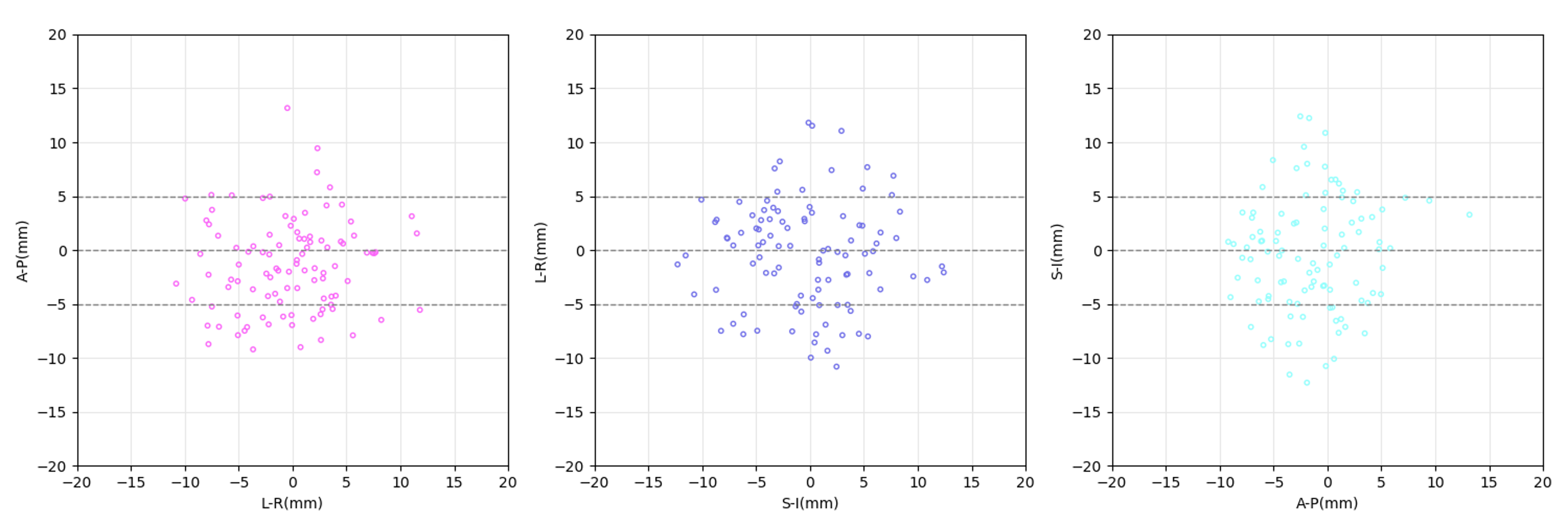}
	\caption{Illustration of the prediction errors observed in the A-P (Anterior-Posterior), L-R (Left-Right), and S-I (Superior-Inferior) directions. The majority of data points exhibit errors within a 5mm range, with the vast majority of points demonstrating errors within a 10mm threshold. It is noteworthy that this is solely achieved through rigid transformation.}
	\label{fig:f7}
\end{figure*}

Our methodology also demonstrated exemplary performance on publicly available cross-modal datasets.As illustrated in Figure \ref{fig:f6}, we present the errors between the predictions generated by our method and the target values in the spatial coordinate system. It is evident that our method exhibits commendable stability in error outcomes and demonstrates a notable clustering effect across various projection planes. As depicted in Figure \ref{fig:f7}, we conducted testing on the Pelvic dataset, and the resultant errors are predominantly within a 5mm range, with the vast majority falling within a 10mm threshold. The primary reason for this relatively larger error can be attributed to our utilization of a translation-based method, which imposes constraints to preserve the structural information of local images, but it may not be conducive to the registration of images with substantial pose disparities. Nevertheless, our method still accomplishes initial pre-alignment even in cases where the images exhibit considerable pose differences, laying a solid foundation for subsequent elastic registration efforts.

\begin{figure*}[htb]
	\centering
	\includegraphics[width=1.0\textwidth,height=0.55\textwidth]{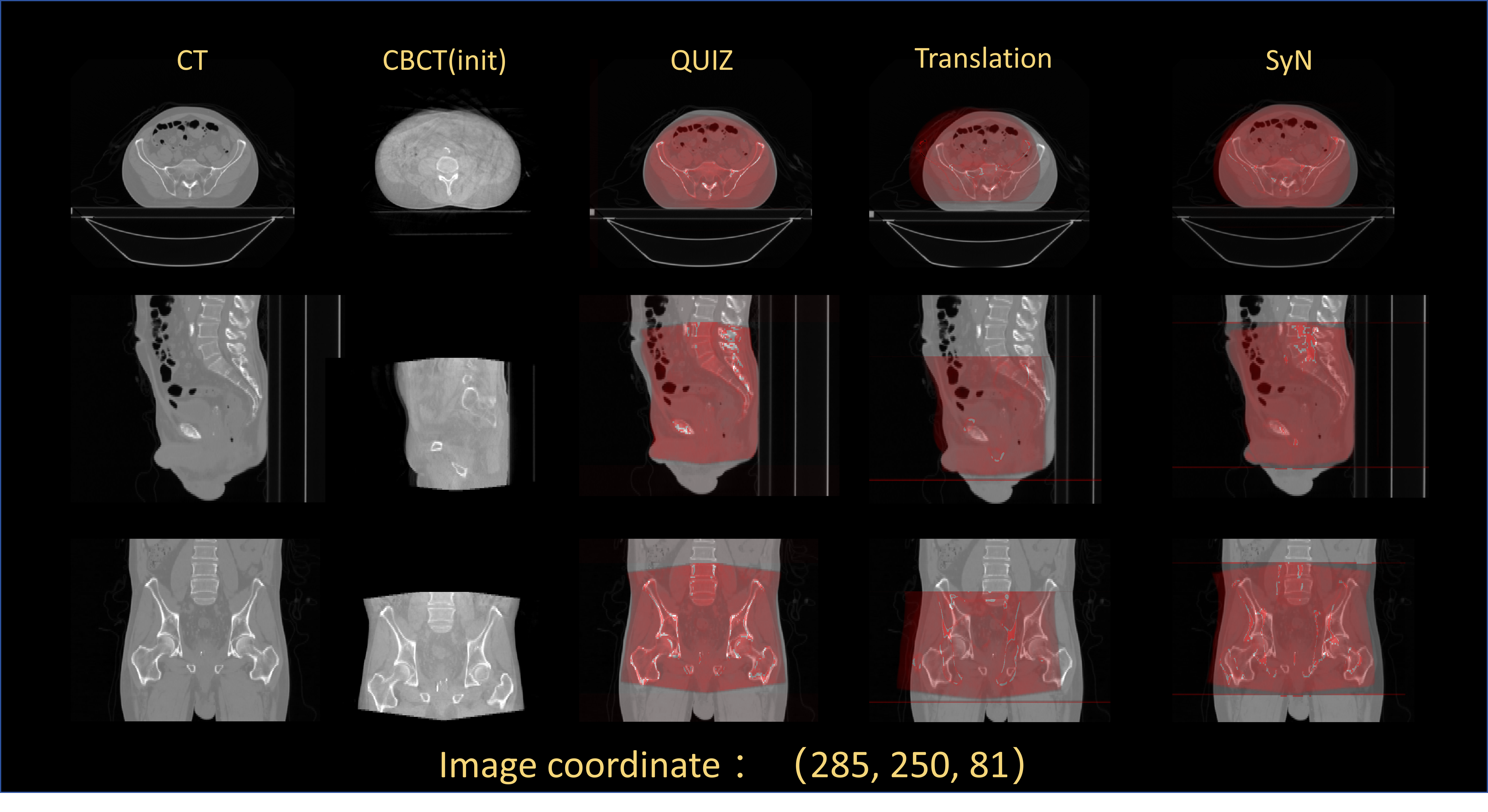}
	\caption{Illustration of cross-modal registration results for a given image pair. The gray area represents the CT image, while the red area signifies the warped CBCT image. It is worth noting that due to the large difference in world coordinates between CT and CBCT, the image coordinates are shown here.}
	\label{fig:f8}
\end{figure*}

\subsection{Limitations of DLIR}

Pre-registration failure can further complicate deformable registration, producing results that are challenging to decipher and may lead to misinterpretations, especially in local-global registration task in our research. Regrettably, numerous current DLIR algorithms predominantly concentrate on the deformable registration module, assuming affine registration \cite{yu2022keymorph}, \cite{mok2022affine}. This assumption could potentially limit the overall performance of these methods in cases where affine registration is not adequate or fails to provide a reliable starting point for deformable registration.

\section{Conclusions}
In this paper, we investigated the challenge of aligning three-dimensional medical images in scenarios involving large deformations. Specifically, we proposed a network explicitly designed to make effective use of semantic information for learning displacements with interpretability, diverging from alternative networks that depend solely on grayscale data for deformation modeling through convolution. This research aims to develop methods that can better account for the specific characteristics of different modalities and applications, while also providing sufficient flexibility to adapt to varying situational requirements. Specifically, we introduced a query-based voxel point matching network that adaptively adjusts the anatomy of interest by introducing query points located at different locations. We also applied our designed method, QUIZ, to the challenging scenario of cervical cancer radiotherapy to showcase its efficacy. This scenario is characterized by high complexity and significant distortions. 

Moreover, our proposed architecture holds potential for application in clinical tasks necessitating precise coordinate positioning. To the best of our knowledge, synchronizing and registering anatomical landmarks or specific meaningful key points across varying spaces or images are commonplace and indispensable during radiotherapy and surgical procedures. The proposed technique stands to benefit challenges that entail identifying correspondences between image pairs predicated on provided 3D points. We believe that the development of such advanced and adaptable methods will enable the medical imaging community to better address the evolving needs and challenges of a variety of clinical use cases.

\section*{Acknowledgement}
This work was supported partly by the grants from the National Natural Science Foundation of China (No.82202954, No.U20A201795, No.U21A20480, No.12126608), Youth Talent Support Programme of Guangdong Provincial Association for Science and Technology (No.SKXRC202224), and the Special Research Assistant Grant Program of the Chinese Academy of Sciences.

\bibliographystyle{elsarticle-num} 
\bibliography{ref}

\begin{thebibliography}{10}
\expandafter\ifx\csname url\endcsname\relax
  \def\url#1{\texttt{#1}}\fi
\expandafter\ifx\csname urlprefix\endcsname\relax\def\urlprefix{URL }\fi
\expandafter\ifx\csname href\endcsname\relax
  \def\href#1#2{#2} \def\path#1{#1}\fi

\bibitem{haskins2020deep}
G.~Haskins, U.~Kruger, P.~Yan, Deep learning in medical image registration: a survey, Machine Vision and Applications 31 (2020) 1--18.

\bibitem{lesterSurveyHierarchicalNonlinear1999}
H.~Lester, S.~R. Arridge, A survey of hierarchical non-linear medical image registration, Pattern Recognition 32~(1) (1999) 129--149.

\bibitem{sengupta2022survey}
D.~Sengupta, P.~Gupta, A.~Biswas, A survey on mutual information based medical image registration algorithms, Neurocomputing 486 (2022) 174--188.

\bibitem{shenImageRegistrationLocal2007}
D.~Shen, Image registration by local histogram matching, Pattern Recognition 40~(4) (2007) 1161--1172.

\bibitem{zheng2021progressive}
Z.~Zheng, W.~Cao, Z.~He, Y.~Luo, Progressive anatomically constrained deep neural network for 3d deformable medical image registration, Neurocomputing 465 (2021) 417--427.

\bibitem{vercauteren2009diffeomorphic}
T.~Vercauteren, X.~Pennec, A.~Perchant, N.~Ayache, Diffeomorphic demons: Efficient non-parametric image registration, NeuroImage 45~(1) (2009) S61--S72.

\bibitem{LORENZI2013470}
M.~Lorenzi, N.~Ayache, G.~Frisoni, X.~Pennec, Lcc-demons: A robust and accurate symmetric diffeomorphic registration algorithm, NeuroImage 81 (2013) 470--483.

\bibitem{1175091}
D.~Shen, C.~Davatzikos, Hammer: hierarchical attribute matching mechanism for elastic registration, IEEE Transactions on Medical Imaging 21~(11) (2002) 1421--1439.

\bibitem{SHEN2009954}
D.~Shen, Fast image registration by hierarchical soft correspondence detection, Pattern Recognition 42~(5) (2009) 954--961.

\bibitem{klein2009elastix}
S.~Klein, M.~Staring, K.~Murphy, M.~A. Viergever, J.~P. Pluim, Elastix: a toolbox for intensity-based medical image registration, IEEE transactions on medical imaging 29~(1) (2009) 196--205.

\bibitem{avants2008symmetric}
B.~B. Avants, C.~L. Epstein, M.~Grossman, J.~C. Gee, Symmetric diffeomorphic image registration with cross-correlation: evaluating automated labeling of elderly and neurodegenerative brain, Medical image analysis 12~(1) (2008) 26--41.

\bibitem{beg2005computing}
M.~F. Beg, M.~I. Miller, A.~Trouv{\'e}, L.~Younes, Computing large deformation metric mappings via geodesic flows of diffeomorphisms, International journal of computer vision 61 (2005) 139--157.

\bibitem{fanBIRNet}
J.~Fan, X.~Cao, P.~Yap, D.~Shen, Birnet: Brain image registration using dual-supervised fully convolutional networks, Medical Image Anal. (2019).

\bibitem{MSnet}
D.~Wei, L.~Zhang, Z.~Wu, X.~Cao, G.~Li, D.~Shen, Q.~Wang, Deep morphological simplification network (ms-net) for guided registration of brain magnetic resonance images, Pattern Recognition 100 (2020) 107171.

\bibitem{RecursiveC}
S.~Zhao, Y.~Dong, E.~Chang, Y.~Xu, Recursive cascaded networks for unsupervised medical image registration, in: IEEE/CVF International Conference on Computer Vision (ICCV), 2019, pp. 10599--10609.

\bibitem{abbas2022}
S.~Abbasi, M.~Tavakoli, H.~R. Boveiri, M.~A. Mosleh~Shirazi, R.~Khayami, H.~Khorasani, R.~Javidan, A.~Mehdizadeh, Medical image registration using unsupervised deep neural network: A scoping literature review, Biomedical Signal Processing and Control 73 (2022) 103444.

\bibitem{7892934}
J.~Rühaak, T.~Polzin, S.~Heldmann, I.~J.~A. Simpson, H.~Handels, J.~Modersitzki, M.~P. Heinrich, Estimation of large motion in lung ct by integrating regularized keypoint correspondences into dense deformable registration, IEEE Transactions on Medical Imaging 36~(8) (2017) 1746--1757.

\bibitem{hansen2020tackling}
L.~Hansen, M.~P. Heinrich, Tackling the problem of large deformations in deep learning based medical image registration using displacement embeddings, in: Medical Imaging with Deep Learning, 2020.

\bibitem{Heinrich2019ClosingTG}
M.~P. Heinrich, Closing the gap between deep and conventional image registration using probabilistic dense displacement networks, in: International Conference on Medical Image Computing and Computer-Assisted Intervention, 2019.

\bibitem{Ma2021DeepLA}
C.~Ma, J.~Zhou, X.~ting Xu, J.~Guo, M.~fei Han, Y.~Gao, H.~Du, J.~N. Stahl, J.~S. Maltz, Deep learning‐based auto‐segmentation of clinical target volumes for radiotherapy treatment of cervical cancer, Journal of Applied Clinical Medical Physics 23 (2021).

\bibitem{vordermark2016radiotherapy}
D.~Vordermark, Radiotherapy of cervical cancer, Oncology research and treatment 39~(9) (2016) 516--520.

\bibitem{swamidas2020image}
J.~Swamidas, C.~Kirisits, M.~De~Brabandere, T.~P. Hellebust, F.-A. Siebert, K.~Tanderup, Image registration, contour propagation and dose accumulation of external beam and brachytherapy in gynecological radiotherapy, Radiotherapy and Oncology 143 (2020) 1--11.

\bibitem{Rigaud2019DeformableIR}
B.~Rigaud, B.~Rigaud, A.~H. Klopp, S.~S. Vedam, A.~M. Venkatesan, N.~Taku, A.~Simon, P.~Haigron, R.~de~Crevoisier, K.~Brock, G.~Cazoulat, Deformable image registration for dose mapping between external beam radiotherapy and brachytherapy images of cervical cancer, Physics in Medicine \& Biology 64 (2019).

\bibitem{Bondar2012IndividualizedNA}
M.~L. Bondar, M.~S. Hoogeman, J.~W.~M. Mens, S.~Quint, R.~Ahmad, G.~Dhawtal, B.~J. Heijmen, Individualized nonadaptive and online-adaptive intensity-modulated radiotherapy treatment strategies for cervical cancer patients based on pretreatment acquired variable bladder filling computed tomography scans., International journal of radiation oncology, biology, physics 83 5 (2012) 1617--23.

\bibitem{Bondar2010ASN}
L.~Bondar, M.~S. Hoogeman, E.~M.~V. Osorio, B.~J. Heijmen, A symmetric nonrigid registration method to handle large organ deformations in cervical cancer patients., Medical physics 37 7 (2010) 3760--72.

\bibitem{r11}
T.~C. Mok, A.~Chung, Fast symmetric diffeomorphic image registration with convolutional neural networks, in: Proceedings of the IEEE/CVF conference on computer vision and pattern recognition, 2020, pp. 4644--4653.

\bibitem{r12}
A.~V. Dalca, G.~Balakrishnan, J.~Guttag, M.~R. Sabuncu, Unsupervised learning for fast probabilistic diffeomorphic registration, in: International Conference on Medical Image Computing and Computer-Assisted Intervention, Springer, 2018, pp. 729--738.

\bibitem{r13}
G.~Balakrishnan, A.~Zhao, M.~R. Sabuncu, J.~Guttag, A.~V. Dalca, An unsupervised learning model for deformable medical image registration, in: Proceedings of the IEEE conference on computer vision and pattern recognition, 2018, pp. 9252--9260.

\bibitem{r14}
G.~Balakrishnan, A.~Zhao, M.~R. Sabuncu, J.~Guttag, A.~V. Dalca, Voxelmorph: a learning framework for deformable medical image registration, IEEE transactions on medical imaging 38~(8) (2019) 1788--1800.

\bibitem{t04}
J.~Chen, E.~C. Frey, Y.~He, W.~P. Segars, Y.~Li, Y.~Du, Transmorph: Transformer for unsupervised medical image registration, Medical image analysis 82 (2022) 102615.

\bibitem{ants}
B.~B. Avants, N.~Tustison, G.~Song, et~al., Advanced normalization tools (ants), Insight j 2~(365) (2009) 1--35.

\bibitem{AYorke2021QualityAO}
A.~A. Yorke, G.~C. McDonald, D.~Solis, T.~Guerrero, Quality assurance of image registration using combinatorial rigid registration optimization (corro), Cancer Research and Cellular Therapeutics (2021).

\bibitem{ghose2015review}
S.~Ghose, L.~Holloway, K.~Lim, P.~Chan, J.~Veera, S.~K. Vinod, G.~Liney, P.~B. Greer, J.~Dowling, A review of segmentation and deformable registration methods applied to adaptive cervical cancer radiation therapy treatment planning, Artificial intelligence in medicine 64~(2) (2015) 75--87.

\bibitem{yu2022keymorph}
E.~M. Yu, A.~Q. Wang, A.~V. Dalca, M.~R. Sabuncu, Keymorph: Robust multi-modal affine registration via unsupervised keypoint detection, in: Medical Imaging with Deep Learning, 2022.

\bibitem{mok2022affine}
T.~C. Mok, A.~Chung, Affine medical image registration with coarse-to-fine vision transformer, in: Proceedings of the IEEE/CVF Conference on Computer Vision and Pattern Recognition, 2022, pp. 20835--20844.

\end{thebibliography}

\end{document}